\documentclass[10pt,twocolumn,letterpaper]{article}

\usepackage{cvpr} 
\usepackage{times}
\usepackage{epsfig}
\usepackage{graphicx}
\usepackage{amsmath}
\usepackage{amssymb}
\usepackage{booktabs}
\usepackage{pgfplots}
\pgfplotsset{compat=1.18}
\usepackage{xcolor}
\colorlet{red}{black}
\usepackage{microtype}
\usepackage{todonotes}
\usepackage{colortbl}          
\usepackage{float}             
\usepackage{tabularx}
\newcolumntype{R}{>{\raggedleft\arraybackslash}X}  
\graphicspath{{./}{res/}}
\definecolor{itaI}{RGB}{253,236,219}
\definecolor{itaII}{RGB}{243,205,165}
\definecolor{itaIII}{RGB}{215,165,110}
\definecolor{itaIV}{RGB}{172,118,65}
\definecolor{itaV}{RGB}{112,65,30}
\definecolor{itaVI}{RGB}{52,25,8}

\begin{document}

\title{Dorsal Hand Images for Immersive (XR) and Privacy-preserving\\ Age Assurance and 
Child Safety}

\author{Riccardo Bovo \quad George Loukas \quad Josh P. Davis\\
University of Greenwich, London, UK\\
{\tt\small r.bovo@gre.ac.uk}
}

\maketitle
\thispagestyle{empty}

\begin{abstract}
Ensuring that Extended Reality (XR) environments are age-appropriate is an important regulatory and safety challenge. However, current age assurance operates only at registration and cannot verify the age of the active user during a session. Face-based approaches, the dominant solution in social media and adult platforms, are impractical in XR, because they require removing the headset and taking a self-captured image, often on a mobile app. This both breaks immersion and introduces the privacy risk of sharing face pictures with third parties, which leaves XR platforms without a viable path to continuous, in-session and privacy-preserving age assurance. We propose the dorsal part of the hand as an alternative to the face, by exploiting the egocentric cameras that XR headsets inherently and naturally use to capture gesture interactions. To evaluate this, we collect an age- and sex-stratified, ethnodiverse dataset of 436 participants spanning the minor--adult boundary, captured under unconstrained lighting and orientation conditions. To characterise what is achievable with off-the-shelf methods at the minor--adult boundary, we evaluate standard neural network architectures for age assurance at the legally critical 18-year threshold. Analysis confirms performance is robust to skin-tone variation. On this dataset, the challenge-31 operating point achieves zero minor admission, making the system a viable first-stage filter for age assurance. These findings position dorsal hand morphometrics as an effective and more privacy-preserving biometric modality for in-session age assurance in XR.
\end{abstract}

\section{Introduction}
\label{sec:intro}

Extended Reality (XR) platforms are used by millions of children for interaction and socialisation. In the US, one in five teenagers own an XR device\footnote{https://xra.org/new-xra-survey-finds-one-in-five-teens-own-a-virtual-reality-headset}. In the UK, nearly one in two children have used XR, and around 1.7 million households own a device, a figure projected to double by 2028~\cite{ofcom2024techtracker,idc2024growth}. 
At the same time, it is widely recognised that XR introduces amplified risks for children, including harassment, grooming, and exposure to harmful content due to its immersive nature~\cite{reed2023kids,baldry2024embodied}. Supervision is limited as parents cannot observe in-headset experiences, and the lack of enforced authentication further increases vulnerability~\cite{marks2025amxra,odudu2024technological,noah2022security,jin2024your}. These concerns are consistently highlighted by parents~\cite{schmuecker2024democratizing,hourcade2024understanding}, but are also a major regulatory challenge. Age assurance requirements are being introduced globally, including the European Union, the United Kingdom, Australia, and several U.S. states~\cite{onlinesafetyact2023,australiaAV2024,ofcom2024ageassurance,Rashid2024}. 
The standard age assurance approach for regulatory compliance is to adopt the face-based age estimation approaches designed for social media and mobile phones, but these are highly impractical. \textcolor{red}{This is an architectural constraint: XR pass-through cameras face outward towards the environment and are physically incapable of imaging the wearer's face during a session.} They require removing the headset, using a third party app and uploading a self-captured (selfie) image. Beyond the high privacy risks, which have already been evidenced by reported data breaches\footnote{https://www.theguardian.com/media/2025/oct/09/hack-age-verification-firm-discord-users-id-photos}, removing the headset breaks immersion and limits the frequency of checks needed to mitigate account sharing risks~\cite{2219713826b749669e991138e03e9c8a,jin2024your,baldry2024embodied}. \textcolor{red}{Partial face capture below the headset rim would substantially degrade age estimation accuracy~\cite{tanveer2025overcoming}.}
To overcome these challenges, we propose to leverage XR's built-in front-facing cameras, inferring age from \emph{hand morphometrics}. XR headsets inherently capture high-fidelity RGB and depth data of hands within interaction range~\cite{hu2024apple}.
Prior work in medical dermatology has identified a number of characteristic dorsal hand features that vary systematically with age~\cite{jakubietz2008ageing,lopez2013hallmarks}, including \emph{wrinkle density, melanin distribution, dorsal vein prominence, and dorsal hair growth}. Building on these anatomical observations, recent work by Georgievskaya et al.~\cite{Georgievskaya2024} demonstrates that chronological age can be inferred from dorsal hand images with accuracy approaching that of face-based methods, \textcolor{red}{a finding on adult cohorts that does not extend to the minor--adult boundary: below the late twenties, the characteristic aging signals are dominated by developmental growth effects rather than biological aging, and only become consistent markers thereafter~\cite{Flament2019ClinicalAges}.}
However, prior work focuses on adult populations, leaving discrimination at the minor--adult boundary an open question. This gap is compounded by two further limitations: existing studies are conducted under controlled laboratory conditions with consistent lighting and fixed camera-hand positions whose results may not transfer to real-world conditions~\cite{Georgievskaya2024}; and large-scale hand image datasets, despite containing dorsal views and demographic metadata, are designed for biometric or gesture recognition, lacking  age stratification and ethnodiverse representation necessary for age inference~\cite{Georgievskaya2024,Afifi2018,Nuzhdin2024}.
We address these gaps by collecting an ethnodiverse, age- and sex-stratified dataset of 436 participants under unconstrained conditions representative of XR use, and evaluating neural networks for age assurance.

\begin{figure}[t]
  \centering
  \includegraphics[width=\columnwidth]{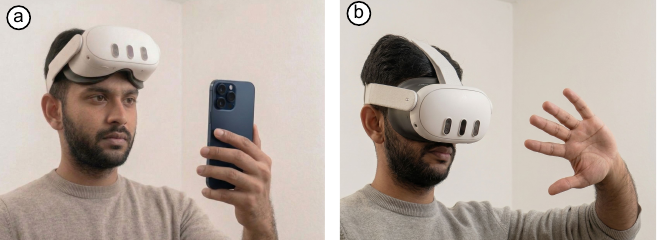}
  \caption{Two age-assurance modalities in XR.
    \textbf{(a)}~Face-based verification requires the user to lift the headset and present their face to a mobile camera: an out-of-band step that breaks immersion.
    \textbf{(b)}~Dorsal hand inference uses the headset's own pass-through RGB sensors to capture the hand during normal interaction, requiring no interruption to the XR session.}
  \label{fig:outofband}
\end{figure}

\section{Related Work}
\label{sec:related}

\subsection{Dermatological Studies on Dorsal Hand Aging}
\label{sec:related:dermatological}

The dorsal hand exhibits a well-characterized aging signature spanning multiple independent features such as wrinkle pattern and density, visible veins, age spots, volume loss, and biophysical properties including skin roughness and chromophore distribution, each progressively correlated with chronological age~\cite{jakubietz2008ageing,Messaraa2019}. Crucially, these features are predominantly visible in standard photography when the dorsal hand fills the frame, making hand images a viable substrate for non-contact age estimation without specialized imaging hardware. However dermatological studies also report that visible signs accumulate non-uniformly across the adult lifespan, with onset later than equivalent facial changes~\cite{Flament2019ClinicalAges}. This raises a particular concern at the minor-to-adult boundary: the visual discriminability of hands just below and just above 18 years of age remains largely uncharacteristic, making it unclear whether photographic hand images can reliably support age-of-majority determination in this critical range. Aging rate also differs substantially across populations: Caucasian and East Asian cohorts exhibit markedly different timelines for the onset of roughness and pigmentation changes~\cite{Messaraa2019}, making demographic diversity a prerequisite for any dataset intended for generalizable inference. 

\subsubsection{High-Resolution Imaging Approaches}

A number of high-resolution and magnified-imaging methods provide quantitative validation that the aging features identified above (i.e., texture, wrinkle microstructure, and epidermal composition) carry measurable, age-discriminative signal: hyperspectral analysis of dorsal hand texture shows that gray-level co-occurrence features correlate with the aging process~\cite{Calin2017AnProcess}; wrinkle-based pipelines extract length, width, and depth descriptors for machine-learning-based age estimation~\cite{kim2009wrinkle}; and epidermal image processing yields chronological age predictors~\cite{Tatsumi1999EstimationProcessing,Tanaka2008QuantitativeAnalysis}. Crucially, these studies rely on magnified or specialised hardware rather than standard RGB photography; leaving open the question of whether equivalent discriminative information survives in consumer-grade images.

\begin{figure*}[t]
  \centering
  \includegraphics[width=\textwidth]{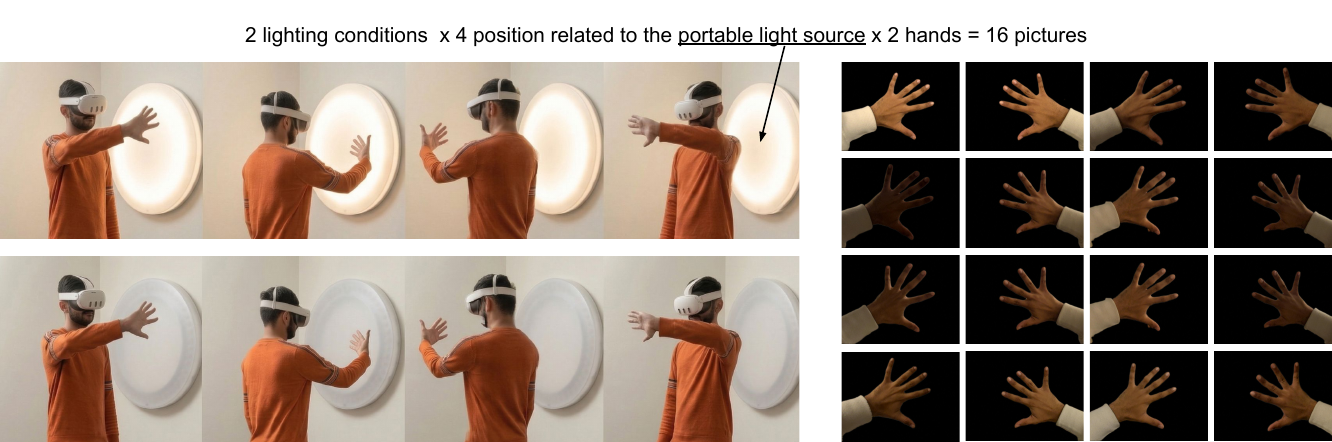}
  \caption{Overview of the image capturing procedure. Each session comprised a protocol of 16 captures per participant, systematically varying users position relative to light source and two intensity levels to yield eight distinct lighting conditions, with each hand photographed once per condition. This photometric diversity was designed to yield a dataset ecologically representative of consumer-grade capture conditions, contrasting with the controlled clinical photography typical of dermatological studies~\cite{Georgievskaya2024,Messaraa2019}.}
  \label{fig:datacollection}
\end{figure*}

\begin{figure*}[t]
  \centering
  \includegraphics[width=\textwidth]{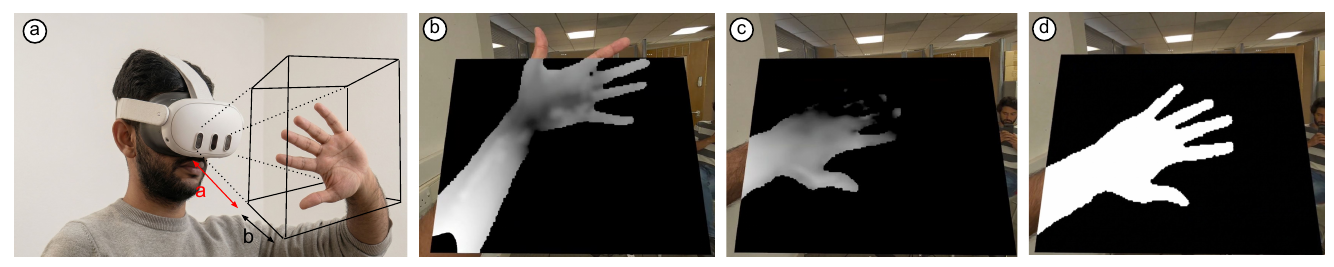}
  \caption{Capture frustum and depth-shader visual feedback used to guide hand positioning during data collection.
    \textbf{(a)}~The capture frustum defines the 3D volume within which the hand must be placed for a valid capture. Dimension $a$ denotes the target distance from the camera (30\,cm) and $b$ the acceptable depth tolerance ($\pm$10\,cm).
    \textbf{(b)}~Hand positioned outside the lateral boundaries of the frustum: the depth shader is dark, indicating no valid capture.
    \textbf{(c)}~Hand within the frustum laterally but too far from the sensor: the shader renders partially dark, prompting the participant to move closer.
    \textbf{(d)}~Hand correctly placed at the target depth within the frustum: the shader renders fully white, confirming a valid capture is in progress.}
  \label{fig:stimuli}
\end{figure*}

\subsection{Computer Vision Approaches}
\label{sec:related:cv}

There are several studies that implement age inference from dorsal hand images~\cite{Baisa2024JointLearning,Abderrahmane2020HandCNN-GRU,Wen2020HandYears,rezasoltani2024multi}, all focused on person identification and gender classification, treating age as an auxiliary classification problem addressed simply by identifying the user, whose data is naturally in the training dataset. This setup inflates reported age performance by allowing models to exploit subject-specific appearance cues rather than learning age-discriminative features that generalize to unseen individuals. Furthermore none of these studies included minor participants. A study that differs is Georgievskaya et al.~\cite{Georgievskaya2024} which collect a purpose-built dataset of Indian females aged 18+ and train CNNs to infer chronological age from dorsal hand images, reporting a mean absolute error (MAE) of 4.7 years approaching face-based accuracy. \textcolor{red}{However, the adult-only cohort cannot address the minor--adult boundary~\cite{Flament2019ClinicalAges}, cross-demographic generalisation, or robustness to unconstrained capture conditions.}

\subsection{Dorsal Hand Datasets with Age Labels}
\label{sec:related:datasets}

The closest public antecedent is 11K Hands~\cite{Afifi2018}: 11,076 RGB images from 190 subjects, yet 96\% of participants are aged 20--29, making it useful for biometric identification but unrepresentative for age assurance. HaGRIDv2~\cite{Nuzhdin2024} is larger (30k images, 7,303 subjects) and better distributed, but its age labels are automatically inferred rather than verified, and neither includes minors. No existing corpus provides age labels, minor--adult coverage, and demographic diversity needed to study the safety-critical , a gap the dataset introduced here is designed to fill.

\begin{figure*}[t]
  \centering
  \includegraphics[width=\textwidth]{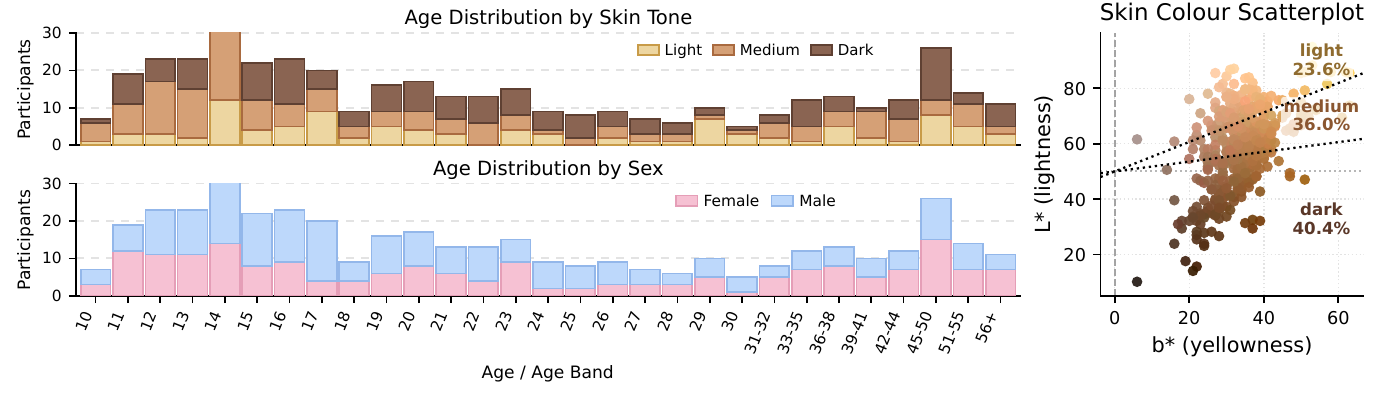}
\caption{(\textit{a})~Age-band composition of the patch cohort by merged skin-tone group
          (light = ITA I--III, medium = IV, dark = V--VII; $n=428$).
          Individual ages are shown from 10--30 years, with grouped bands for ages 31+.
          (\textit{b})~Age-band distribution of the same participants by sex ($n=428$).
          (\textit{c})~Per-participant CIE $L^*$--$b^*$ skin-colour scatterplot for the patch cohort,
          with points coloured by measured patch RGB. Dotted lines mark the ITA boundaries
          between the merged light, medium, and dark groups, and the in-panel labels report
          the proportion of participants in each group.}

  \label{fig:demographics}
\end{figure*}

\section{Dataset}

Existing dorsal hand image datasets and studies lack the characteristics necessary to study the minor--adult boundary under real-world conditions (Sec.~\ref{sec:intro}). To address this, we collected a purpose-built corpus of dorsal hand images from \textcolor{red}{436} participants spanning ages 10--67, captured under the unconstrained lighting and hand-orientation conditions of consumer XR headsets. The dataset is stratified by age and sex to ensure representative coverage near the legally critical 18-year threshold, and spans the full skin-tone spectrum to support fairness analysis. Three parallel channels were used to achieve demographic coverage across the target age range of 10--67 years. A recruitment company was contracted to recruit children aged 10--17 and their guardians.  Lab sessions ran at the University of Greenwich during October--November 2025.  Eligible families received a \pounds60 voucher (\pounds80 for two children). Furthermore, the team visited secondary schools across London between December 2025 and January 2026. In parallel, University of Greenwich staff and undergraduate and postgraduate students were recruited via internal communications to fill adult age-band quotas. Inclusion required (a) age 10 or over, (b) no conditions affecting hand mobility, and (c) ability to tolerate a short VR session. Those reporting epilepsy or motion sickness were excluded.  The final dataset comprises \textcolor{red}{436 participants, yielding 6,790 dorsal hand images (mode 16 per participant, with some deviation due to missing captures)}. Fig.~\ref{fig:demographics} visualizes the full per-band breakdown using the age bands defined. The minor cohort (10--17) constitutes \textcolor{red}{45.9\%} ($n=200$) and adults \textcolor{red}{54.1\%} (\textcolor{red}{$n=236$}).  The sex split is 56.6\% male and 43.4\% female. Stratification targets were set \emph{a priori} for each band to ensure balanced year-of-age and sex cell coverage.

\subsection{XR Equipment and Application}
\label{sec:equipment}
The Meta Quest~3 was selected for its wide consumer adoption and ease of development. Raw camera frames were accessed via the Meta Passthrough Camera API (PCA)\footnote{\url{https://developers.meta.com/horizon/documentation/native/android/pca-native-overview/}} at a resolution of 1280$\times$960\,px. \textcolor{red}{Background pixels were removed using the headset's per-pixel depth stream prior to storage; no background cues are therefore available to the models, and backgrounds are not included in the  dataset.} The researchers monitored what participants could see via scrcpy\footnote{\url{https://scrcpy.org/}} and provided verbal guidance where needed. A key design decision was to capture hand images under \emph{unconstrained conditions}, in contrast to prior datasets. Georgievskaya et al.~\cite{Georgievskaya2024} stabilised hands using a physical rest at a fixed distance from the camera under controlled cross-polarised studio lighting. The 11K Hands dataset~\cite{Afifi2018} was collected with hands placed flat on a table at a fixed camera distance and under fixed lighting. In our protocol, participants were instead asked to raise and hold their hand freely within the capture frustum, guided only by the depth-shader visual feedback (Fig.~\ref{fig:stimuli}). This introduces variability in hand position, depth, and in-plane rotation, while the frustum geometry and the visual feedback cue jointly ensure that the dorsal surface remains approximately perpendicular to the camera sensors and within the valid sensing range. The resulting images are therefore representative of the kind of in-session captures that would occur in a real consumer XR deployment. Given the sensitivity of biometric imagery collected from minor participants, the dataset will not be made publicly available or shared outside the research team.

\subsection{Procedure}
\label{sec:procedure}

Upon arrival, participants completed a pre-screening questionnaire and signed an information sheet and consent form. The VR session lasted approximately five minutes and comprised 16 captures distributed across eight conditions ($2~\text{lighting intensities} \times 4~\text{lighting origins}$); see Fig.~\ref{fig:datacollection}. In each condition, the participant was guided by on-screen audio-visual cues to position their hand within the capture frustum. Four frames were captured for both the dorsal and palmar surfaces of each hand per condition. The session concluded with a debrief and a gift card.

\subsection{Skin Tone Diversity}

Skin tone was quantified using the Individual Typology Angle
(ITA)~\cite{Chardon1991}, computed from the mean CIE~$L^*$ and $b^*$ values of a $50\times50$\,px dorsal-hand patch extracted from a single capture per participant at the position of strongest dorsal illumination.  ITA is defined as $\text{ITA} = \arctan\!\bigl((L^*-50)/b^*\bigr)\times(180/\pi)$,
with larger values indicating lighter skin. As images were captured under unconstrained lighting conditions without color calibration targets, these measurements should be interpreted as a best-effort proxy rather than precise colorimetric estimates, and are used primarily to characterize relative skin-tone diversity within the dataset.
Fig.~\ref{fig:demographics} shows the grade distribution and per-participant $L^*$--$b^*$ positions for the 428 participants with  successfully extracted and usable dorsal-hand patches. The dataset spans the full ITA spectrum (grades~I--VII), with 40.4\% of participants classified as dark, 36.0\% as medium, and 23.6\% as light skin tone.
\subsection{Ethics}
\label{sec:ethics}

The study received ethics approval from the authors' institutional Ethics Committee; all procedures complied with the UK Data Protection Act 2018 and UK GDPR. Adults provided written informed consent; for participants under 18, guardian consent and child assent were obtained prior to any data capture. Participants were assigned randomly generated unique identifiers (UIDs) used exclusively in data files, with age and sex stored in a separate record joined by UID only; a participant-chosen PIN was held in a secure pseudonymisation table, and participants could request deletion within a 14-day window, after which the PIN--UID link was destroyed. All child-facing researchers have undergone enhanced background checks and hold current certification; a school staff member was present throughout every school visit and no researcher was ever alone with a child.

\section{Methods and Experiment}
\label{sec:experiments}

To characterise what is achievable with standard methods on this new problem, we evaluate four neural architectures spanning a range of capacity and efficiency trade-offs on two tasks: continuous age regression and binary adult/minor classification (the age gate). All models are trained and evaluated under identical conditions using participant-level stratified 5-fold cross-validation,
ensuring reliable estimates of generalization to unseen participants. 
Each backbone represents a distinct design point.
\textbf{ResNet-50}~\cite{he2016deep} at $224\times224$ serves as a canonical CNN baseline familiar to the biometric and medical-imaging community. \textbf{EfficientNetV2-S}~\cite{tan2021efficientnetv2} at $384\times384$ is the primary CNN candidate: its compound-scaled architecture preserves the fine-grained texture cues (wrinkle density, melanin distribution) that age estimation relies on. 
 \textbf{Swin Transformer V2-B}~\cite{liu2022swin} at $384\times384$ provides a hierarchical attention baseline; its shifted-window self-attention is better suited to spatially localized hand texture than global-attention ViTs, and the V2 design explicitly targets high-resolution generalization. \textbf{MobileNetV3-Large}~\cite{howard2019searching} at $224\times224$ probes the accuracy--efficiency frontier relevant to on-device XR inference. Input resolutions follow each backbone's canonical ImageNet pretraining configuration, maximizing transfer from pretrained weights given the relatively modest dataset size.

\subsection{Data Splits}
\label{sec:splits}

Splits are defined at the \emph{participant} level to prevent identity leakage. A stratified 15\% held-out test set is drawn first (age bins of 2 years up to 50, 5 years thereafter, jointly stratified on age and skin-tone group); the remaining 85\% is split into 5-fold cross-validation with the same scheme, yielding approximately 68\%\,/\,17\% train/val per fold.

\subsection{Data Augmentation}
\label{sec:augmentation}

Because hand orientation during XR interaction is unconstrained, the training pipeline applies spatial augmentation: random rotation over the full $[-360^{\circ}, 360^{\circ}]$ range, random horizontal flip, and random vertical flip, each applied independently. Mild photometric perturbations (brightness and contrast jitter of $\pm10\%$) account for the variable lighting conditions of the eight capture scenarios. At inference, only a centre-crop resize is applied. All images are normalised with ImageNet channel statistics (mean $[0.485, 0.456, 0.406]$, std $[0.229, 0.224, 0.225]$) to match the pre-training distribution of the backbone weights.

\subsection{Training Details}
\label{sec:training}

All models are trained with AdamW~\cite{loshchilov2017decoupled} ($\text{lr}=3\times10^{-4}$, weight decay $10^{-2}$) for up to 40 epochs with early stopping (patience 20 epochs, minimum delta $10^{-3}$) on validation loss.
The batch size is 32; all models minimize the Gaussian NLL objective defined in Eq.~\ref{eq:nll}, and all experiments use a fixed random seed of~42. All models were trained using distributed data-parallel training across 8$\times$ NVIDIA A100-SXM4 (80GB) GPUs on the University of Greenwich HPC cluster.

\subsection{Probabilistic Age Regression}
\label{sec:loss}

Following~\cite{kendall2017uncertainties},  each model outputs a predicted age mean $\mu$ and log-variance $\log\sigma^2$.
Training minimises the Gaussian negative log-likelihood:
\begin{equation}
\mathcal{L} = \tfrac{1}{2}\!\left(\log\sigma^2 + \frac{(y-\mu)^2}{\sigma^2}\right),
\label{eq:nll}
\end{equation}
where $y$ is the ground-truth age and $\sigma^2 = \exp(\log\sigma^2)$, with $\log\sigma^2$ clamped to $[-10, 10]$ for numerical stability.
Jointly optimizing $\mu$ and $\sigma^2$ encourages well-calibrated uncertainty alongside point predictions, which are exploited directly during Age Gate calibration (Sec.~\ref{sec:tau}).
Because the dataset is not uniform across ages, samples from rare age groups would otherwise contribute disproportionately little to the gradient. To correct for this, following~\cite{cui2019class}, each training sample receives an inverse-frequency weight:
\begin{equation}
w(a) = \operatorname{clip}\!\left(\frac{\bigl(\text{count}(a)+\varepsilon\bigr)^{-\gamma}}
                                      {\mathbb{E}_a\!\left[\bigl(\text{count}(a)+\varepsilon\bigr)^{-\gamma}\right]},\;
                              w_{\min},\, w_{\max}\right),
\label{eq:reweight}
\end{equation}
where $\mathbb{E}_a[\cdot]$ is the training-age mean, with $\varepsilon=1$, $\gamma=1$, $w_{\min}=0.25$, $w_{\max}=4.0$. The weight $w(a)$ is applied via a weighted mean reduction inside the NLL loss, so rare ages exert proportionally stronger influence on the gradient without allowing any single age to dominate.

\subsection{Binary Age Gate (Age Assurance)}
\label{sec:tau}

Age assurance reduces age regression to a binary decision (adult/minor): given a dorsal hand image, determine whether the participant is an adult ($\geq$18 years).
We obtain a calibrated probability from the model's Gaussian output:
\begin{equation}
p_{\mathrm{adult}} = P(a \geq 18) = 1 - \Phi\!\left(\frac{18-\mu}{\sigma}\right),
\label{eq:prob}
\end{equation}
where $\Phi$ is the standard normal CDF.
A participant is admitted as an adult if $p_{\mathrm{adult}} \geq \tau$; otherwise they are flagged for age verification. The decision threshold $\tau$ is selected \emph{per model} by balancing the false positive rate (FPR) of falsely accepting minors against the true positive rate (TPR) of correctly accepting adults, following the operating-point framework of~\cite{hanaoka2024face}.

\begin{figure}[t]
  \centering
  \includegraphics[width=\columnwidth]{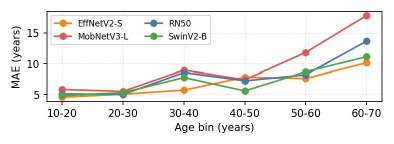}
  \includegraphics[width=\columnwidth]{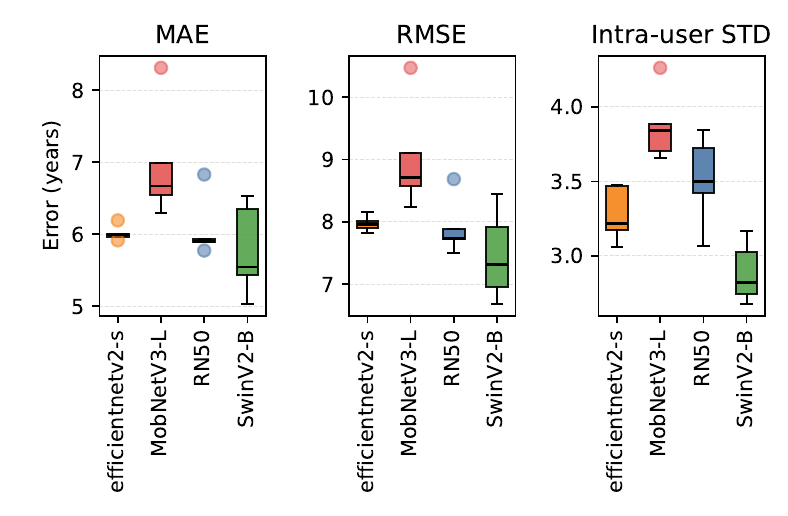}
  \caption{Age regression results across architectures.
    \textbf{Top:} MAE by decade.
    \textbf{Bottom:} Overall MAE, RMSE, and intra-user STD
    (mean\,$\pm$\,std across 5 held-out test folds).}
  \label{fig:age_regression}
\end{figure}

\begin{table}[t]
\centering
\caption{Age regression performance (mean\,$\pm$\,std, 5 folds).}
\label{tab:age_regression}
\small\begin{tabularx}{\columnwidth}{lRRR}
\toprule
Model & MAE (yr) & RMSE (yr) & Intr-usr~STD \\
\midrule
\textbf{SwinV2-B}   &\textbf{5.78\,$\pm$\,0.64} & \textbf{7.47\,$\pm$\,0.72} & \textbf{2.89\,$\pm$\,0.20} \\
EffNetV2-S & 6.01\,$\pm$\,0.11 & 7.97\,$\pm$\,0.13 & 3.28\,$\pm$\,0.19 \\
RN50       & 6.07\,$\pm$\,0.43 & 7.91\,$\pm$\,0.46 & 3.51\,$\pm$\,0.30 \\
MobNetV3 & 6.96\,$\pm$\,0.79 & 9.02\,$\pm$\,0.87 & 3.87\,$\pm$\,0.24 \\
\bottomrule
\end{tabularx}
\end{table}

\section{Results}
\label{sec:results}

We evaluate the four architectures on two tasks: continuous age regression and binary adult/minor classification (age gate). All cross-model comparisons use paired tests across the 5 held-out folds (4 degrees of freedom); omnibus differences are assessed with repeated-measures ANOVA and Friedman test; post-hoc pairwise differences use a paired $t$-test with Holm--Bonferroni correction for the $\binom{4}{2}=6$ comparisons ($\alpha=0.05$).

\subsection{Age Regression}
\label{sec:results:regression}

We report MAE and RMSE as point-accuracy measures, and intra-user STD (16 per-participant captures) as a measure of stability for the age-gate decision. SwinV2-B achieves the lowest MAE ($5.78\pm0.64$\,yr) and RMSE ($7.47\pm0.72$\,yr), followed by EfficientNetV2-S ($6.01\pm0.11$ MAE). MobileNetV3-L performs worst across all metrics. Omnibus tests confirm significant cross-model differences in all three metrics. After Holm correction, pairwise MAE differences are not individually significant; however, the RMSE gap between MobileNetV3-L and SwinV2-B reaches significance ($p{=}0.016$).


\begin{figure}[t]
  \centering
  \begin{minipage}[t]{0.48\columnwidth}
    \centering
    \includegraphics[width=\linewidth]{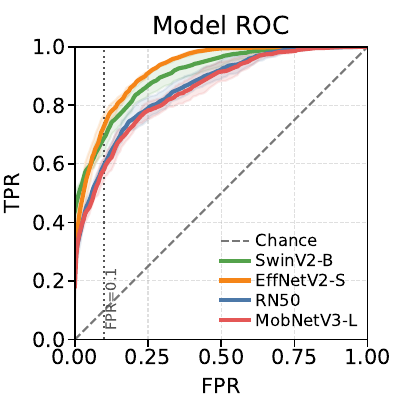}
  \end{minipage}\hfill
  \begin{minipage}[t]{0.48\columnwidth}
    \centering
    \includegraphics[width=\linewidth]{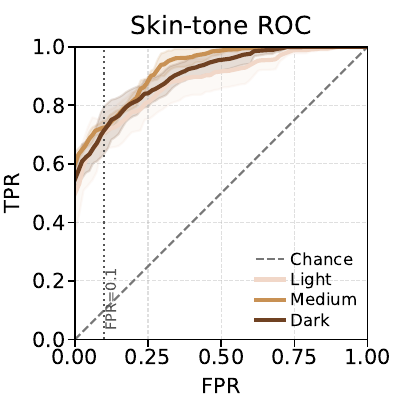}
  \end{minipage}
  \caption{\textbf{(a)}~ROC curves for the binary adult/minor age gate across architectures
    (mean\,$\pm$\,std, 5 held-out test folds). The dotted line marks $\mathrm{FPR}=0.1$.
    \textbf{(b)}~SwinV2-B ROC curves disaggregated by skin-tone group
    (Light\,=\,ITA~I--III; Medium\,=\,IV; Dark\,=\,V--VII).
    Shaded bands show std across 5 folds.}
  \label{fig:roc_age_gate}
\end{figure}

 
\begin{table}[t]
\centering
\caption{Age-gate results.
  \textbf{(A)} Cross-model AUC and pAUC@0.1 (mean\,$\pm$\,std, 5 folds).
  \textbf{(B)} SwinV2-B FPR (minors) and FNR (adults) at three thresholds (mean, 5 folds).}
  \vspace{2pt}
\label{tab:agegate}
\small
\begin{tabularx}{\columnwidth}{XRR}
\multicolumn{3}{l}{\textbf{(A) Cross-model AUC and pAUC@0.1}} \\
\toprule
Model & AUC & pAUC@0.1 \\
\midrule
SwinV2-B & 0.905\,$\pm$\,0.019 & \textbf{0.0587\,$\pm$\,0.0018} \\
EffNetV2-S & \textbf{0.922\,$\pm$\,0.008} & 0.0561\,$\pm$\,0.0057 \\
RN50              & 0.859\,$\pm$\,0.018           & 0.0464\,$\pm$\,0.0019 \\
MobNetV3-L        & 0.851\,$\pm$\,0.018           & 0.0451\,$\pm$\,0.0022 \\
\bottomrule
\end{tabularx}
\vspace{4pt}
{\setlength{\tabcolsep}{1pt}\begin{tabularx}{\columnwidth}{RRRRRRRRRRR}
\multicolumn{11}{l}{\textbf{(B) SwinV2-B FPR and FNR per age buckets}} \\
\toprule
 & \multicolumn{3}{c}{FPR (\%)} & \multicolumn{7}{c}{FNR (\%)} \\
\cmidrule(lr){2-4}\cmidrule(lr){5-11}
$\tau$ & 10\textendash12 & 13\textendash15 & 16\textendash17 & 18\textendash19 & 20\textendash24 & 25\textendash29 & 30\textendash39 & 40\textendash49 & 50+ & All \\
\multicolumn{1}{l}{\scriptsize $n$} & \scriptsize 8x16& \scriptsize 14x16& \scriptsize 7x16& \scriptsize 4x16& \scriptsize 9x16& \scriptsize 7x16& \scriptsize 5x16& \scriptsize 6x16& \scriptsize 6x16 & \\
\midrule
$0.775$ & 5 & 14 & 22 & 50 & 52 & 22 & 29 &  0 & 0 & 26 \\
$0.923$ & 0 &  3 &  6 & 86 & 80 & 72 & 61 &  7 & 0 & 52 \\
$0.960$ & 0 &  0 &  0 & 95 & 88 & 84 & 77 & 25 & 7 & 63 \\
\bottomrule
\end{tabularx}}
\end{table}

\subsection{Age Gate}
\label{sec:results:gate}

The probabilistic outputs ($\mu$, $\sigma$) are reduced to a binary adult/minor decision following Sec.~\ref{sec:tau}. The partial AUC at FPR$\leq$0.1 (pAUC@0.1) summarises discrimination in the safety-relevant low-false-acceptance region, computed as the area under the ROC curve integrated over $[0, 0.1]$ and normalised to $[0, 0.1]$. EfficientNetV2-S achieves the highest overall AUC ($0.922\pm0.008$); however, SwinV2-B achieves the highest pAUC@0.1 ($0.0587\pm0.0018$), the safety-critical metric that summarises discrimination in the low-false-acceptance region. SwinV2-B is therefore selected as the recommended model for age-gate deployment (Fig.~\ref{fig:roc_age_gate}(a)). Its pAUC advantage over RN50 ($\Delta{=}+0.0123$) and MobileNetV3-L ($\Delta{=}+0.0135$) is statistically significant after Holm correction ($p{=}0.0007$ for both).

\subsubsection{Operating Threshold Analysis and Equivalence with NIST Challenge-$T$.}
\label{sec:results:threshold}

Table~\ref{tab:agegate}B reports SwinV2-B held-out FPR and FNR at three decision thresholds $\tau$, aggregated across the five test folds. Because the model outputs a Gaussian $(\mu, \sigma)$, each threshold $\tau$ implies a sample-adaptive effective challenge age:
\begin{equation}
  T_{\text{eff}} = 18 - \sigma \cdot \Phi^{-1}(1 - \tau),
  \label{eq:teff}
\end{equation}
where $\Phi^{-1}$ is the probit function. A participant is admitted when 
$\mu \geq T_{\text{eff}}$, so higher $\tau$ imposes a larger buffer above the legal threshold, directly analogous to the Challenge-$T$ construct used in NIST FATE AEV evaluations~\cite{hanaoka2024fate}, where a system flags anyone whose estimated age falls below $T > 18$ for secondary verification. With $\text{MAE} \approx 0.798\,\bar{\sigma}$ for a well-calibrated Gaussian, SwinV2-B's MAE of $5.78$ years implies $\bar{\sigma} \approx 7.2$ years (assuming Gaussian calibration). The three thresholds mapping is displayed in Table~\ref{tab:challenge_equiv}.

\begin{table}[H]
\centering
\caption{Approximate NIST Challenge-$T$ equivalents for each decision
threshold $\tau$, derived via Eq.~\ref{eq:teff} with
$\bar{\sigma} \approx 7.2$ years (inferred from SwinV2-B MAE\,=\,5.78\,yr under the Gaussian calibration assumption; see text).}
\label{tab:challenge_equiv}
\small\begin{tabularx}{\columnwidth}{lllX}
\toprule
$\tau$ & $\Phi^{-1}(1-\tau)$ & $T_{\text{eff}}$ & NIST analogue \\
\midrule
$0.775$ & $-0.76$ & $23.2$ & \mbox{Challenge-23} \\
$0.923$ & $-1.43$ & $27.9$ & \mbox{Challenge-28} \\
$0.960$ & $-1.75$ & $30.1$ & \mbox{Challenge-31} \\
\bottomrule
\end{tabularx}

\end{table}

Under this calibration assumption, the recommended operating point $\tau{=}0.923$ corresponds approximately to the Challenge-25/28 regime evaluated in NIST FATE AEV~\cite{hanaoka2024fate}, and achieves a 16--17 FPR of 6\%, consistent with the credible system range of 2--10\% reported for top-tier commercial algorithms on high-quality application images~\cite{hanaoka2024fate,biometricupdate2024fate}. 
The strict setting $\tau{=}0.960$ (Challenge-31) reduces minor admission to zero, flagging nearly all adults under 40 (overall adult FNR: 63\%). Unlike fixed Challenge-$T$ systems, the probabilistic formulation in Eq.~\ref{eq:teff} applies a \emph{per-sample} buffer: predictions with higher uncertainty $\sigma$ incur a larger challenge margin, providing conservative treatment of ambiguous cases without a separate calibration step.


\begin{table}[t]
\centering
\caption{SwinV2-B skin-tone analysis.}
  \vspace{2pt}
\label{tab:roc_skin}
\small
\begin{tabularx}{\columnwidth}{XlRR}
\multicolumn{4}{l}{\textbf{(A) Age-gate ROC by skin-tone group (mean\,$\pm$\,std, 5 folds)}} \\
\toprule
Skin tone & $n$/fold & AUC & pAUC@0.1 \\
\midrule
Medium & 164 & 0.926\,$\pm$\,0.008 & 0.0679\,$\pm$\,0.0019 \\
Dark  & 184 & 0.904\,$\pm$\,0.027 & 0.0636\,$\pm$\,0.0054 \\
Light & 108 & 0.881\,$\pm$\,0.042 & 0.0620\,$\pm$\,0.0092 \\
\bottomrule
\end{tabularx}
\vspace{4pt}
\begin{tabularx}{\columnwidth}{XXRRR}
\multicolumn{5}{l}{\textbf{(B) Pairwise pAUC@0.1 comparisons (Holm-corrected, 5 folds)}} \\
\toprule
A & B & $\Delta$pAUC & $p_{\text{Holm}}$ & Sig.\ \\
\midrule
Medium & Dark  & $+$0.0043 & 0.828 & no \\
Medium & Light & $+$0.0059 & 0.828 & no \\
Dark & Light & $+$0.0016 & 0.828 & no \\
\bottomrule
\end{tabularx}
\end{table}

\subsubsection{Skin Tone Analysis}
\label{sec:results:skin}

Skin-tone labels are derived from ITA values computed on dorsal-hand patches without colorimetric calibration, extracted from a single capture at the strongest illumination, representing a best-effort classification. Seven Fitzpatrick grades are collapsed to three groups: Light (I--III), Medium (IV), and Dark (V--VII).
SwinV2-B's age-gate discrimination does not differ significantly across skin-tone groups (Fig.~\ref{fig:roc_age_gate}(b)). Table~\ref{tab:roc_skin} shows pAUC@0.1 values of $0.0679$, $0.0636$, and $0.0620$ for Medium, Dark, and Light, respectively; pairwise Holm-corrected comparisons yield $p{>}0.82$ for all pairs (Table~\ref{tab:roc_skin}B), suggesting that the binary safety boundary is robust to skin-tone variation within this dataset.

\subsection{\textcolor{red}{Multi-Capture Aggregation}}
\label{sec:results:aggregation}

\textcolor{red}{Each session yields 16 captures per participant, enabling score-level aggregation at inference: per-image $p_{\text{adult}}$ scores from $n$ captures are averaged before applying the decision threshold $\tau$, without retraining. Table~\ref{tab:aggregation} reports AUC and age-gate outcomes at $\tau{=}0.923$ for $n\in\{1,2,4\}$. Aggregation yields a Pareto improvement at the recommended operating point: the 16--17-year-old FPR drops from 6\% ($n{=}1$) to 2.2\% ($n{=}2$) to \textbf{0\%} ($n{=}4$), and the total minor FPR falls from 0.9\% to 0\%, while the adult pass rate remains unchanged at $\sim$48\%. AUC gains are consistent across all architectures ($\Delta$AUC$_{n=1\to4}$: $+$0.029 SwinV2-B, $+$0.035 EffNetV2-S, $+$0.040 ResNet-50, $+$0.038 MobileNetV3-L), and participant coverage remains 100\% at $n\leq4$.}

\begin{table}[t]
\centering
\caption{\textcolor{red}{Multi-capture aggregation.}}
\label{tab:aggregation}
\small
\begin{tabularx}{\columnwidth}{lRRR}
\multicolumn{4}{l}{\textbf{(A) AUC by model and aggregation level}} \\
\toprule
Model & AUC ($n{=}1$) & AUC ($n{=}4$) & $\Delta$AUC \\
\midrule
SwinV2-B   & 0.905 & 0.934 & $+$0.029 \\
EffNetV2-S & 0.922 & 0.957 & $+$0.035 \\
ResNet-50  & 0.859 & 0.899 & $+$0.040 \\
MobNetV3-L & 0.851 & 0.889 & $+$0.038 \\
\bottomrule
\end{tabularx}
\vspace{4pt}
\begin{tabularx}{\columnwidth}{lRRR}
\multicolumn{4}{l}{\textbf{(B) SwinV2-B FPR at $\tau{=}0.923$}} \\
\toprule
 & $n{=}1$ & $n{=}2$ & $n{=}4$ \\
\midrule
16--17 FPR (\%)      & 6.0 & 2.2 & \textbf{0} \\
Total minor FPR (\%) & 0.9 & --  & \textbf{0} \\
Adult PR (\%)        & 48  & 48  & 48         \\
\bottomrule
\end{tabularx}
\end{table}
 
\section{Discussion}
\label{sec:discussion}

The results demonstrate that dorsal hand morphology carries sufficient discriminative signal for minor--adult age assurance under unconstrained, ecologically valid conditions, advancing a modality that prior work had only evaluated on adult cohorts under lab settings. The most immediate deployment context is immersive XR, where the dorsal hand is naturally visible to headset pass-through cameras. Current face-based age verification approaches are out-of-band and high-friction, interrupting the XR session and requiring users to present their face to a third-party verifier. Dorsal hand inference offers a passive in-session alternative: the probabilistic formulation maps directly onto the NIST FATE Challenge-$T$ framework~\cite{hanaoka2024fate}, with the recommended operating point ($\tau{=}0.923$, \mbox{Challenge-28} regime) admitting up to 6\% of 16--17-year-olds from a single capture, \textcolor{red}{a figure that multi-capture aggregation reduces to 0\% at $n{=}4$ (Sec.~\ref{sec:results:aggregation}), while the adult pass rate remains 48\%}, and the strict setting ($\tau{=}0.960$, \mbox{Challenge-31}) achieving zero minor admission. At the strict threshold, the system acts as a first-stage conservative filter, escalating borderline cases to higher-friction document-based verification~\cite{ofcom2024ageassurance} rather than replacing it entirely; at more permissive thresholds, it supports adaptive in-session safety controls (such as automatically muting voice chat or restricting social features for users classified as minors) without interrupting user experience.

Direct numerical comparison with the only prior dorsal hand MAE figure of $4.7$ years reported by Georgievskaya et al.~\cite{Georgievskaya2024} is not straightforward: their study was conducted on a single-demographic cohort of adult Indian women, imaged under calibrated cross-polarised lighting with a standardised hand pose, a substantially easier task than the one studied here. Our dataset spans both sexes, the full skin-tone spectrum, a wide array of lighting conditions without colorimetric calibration, and crucially includes the minor adult boundary where no prior dorsal hand study has evaluated performance. The appropriate interpretation of our best MAE of $5.78$ years is therefore not as an under-performance relative to that benchmark, but as a first characterization of the problem under the conditions that actually matter for deployment. 

As accuracy scales with data, the same approach extends to mobile devices, offering a more privacy-preserving alternative to face-based. Dorsal hand images are still biometric data~\cite{gonzalez2024contactless}, but they carry less practical re-identification risk than faces. Faces are everywhere, on social media, CCTV, and personal devices, and backed by massive labelled datasets that a third party could match against. Hands are not, so even if a hand image leaked, there is far less infrastructure available to link it back to an individual. A hand-based system therefore does not expose children's faces to operators or third-party verifiers, making it well suited to contexts where child protection and data minimisation must coexist~\cite{ofcom2024overview,Rashid2024}.

\section{Limitations and Future Work}
\label{sec:future}

Some limitations bound the current findings. The 6\% FPR for 16--17-year-olds at $\tau{=}0.923$ is comparable to top-tier face-based systems on high-quality images~\cite{hanaoka2024fate,biometricupdate2024fate}; this rate is promising for a first system yet below the threshold for unassisted deployment without a secondary verification step. The Challenge-$T$ equivalence in Table~\ref{tab:challenge_equiv} rests on the Gaussian calibration assumption, which should be validated in the region 15--21 where it matters most. The dataset comprises \textcolor{red}{436} participants; performance under greater demographic diversity, and uncontrolled real-world capture remains untested. Finally, ITA is used as a skin-tone proxy but it is uncalibrated here, limiting the precision of the equitability analysis.

Two directions stand out for future work. First, the dorsal hand exhibits well-documented biological aging cues: wrinkle density, dorsal vein prominence, age-spot distribution, volume loss, and skin roughness~\cite{jakubietz2008ageing,Messaraa2019,Flament2019ClinicalAges,Zeljkoviu2021}. Grounding model decisions in these cues via gradient-based saliency maps aligned with expert-annotated feature regions would validate that models exploit biologically meaningful signals rather than artefacts.
Second, several of these cues are inherently three-dimensional: vein prominence, volume loss, and surface micro-roughness manifest as geometric deformations that RGB imaging captures only indirectly through shading. Recent XR headsets, including the Meta Quest~3 used here, Apple Vision Pro, Samsung Galaxy XR and others, provide per-pixel depth alongside RGB at no additional hardware cost~\cite{hu2024apple}, making RGB-D fusion a natural next step for improving discrimination of volume-related cues, particularly near the minor--adult boundary where pigmentation and wrinkle signals are the weakest.

\section{Conclusion}
\label{sec:conclusion}

This work presents the first age-stratified, ethnodiverse dorsal hand age inference study spanning the minor--adult boundary (ages 10+) and the first empirical evidence that dorsal hand RGB images can support safety-critical age assurance. We establish a first benchmark on this dataset, finding that SwinV2-B achieves MAE $5.78$\,yr and pAUC@0.1 $0.0587$ characterising what standard architectures can achieve at the minor--adult boundary. 
The study contributes two firsts to the field: inclusion of minor participants, enabling evaluation at the legally critical 18-year boundary; and ITA-based skin tone characterisation spanning the full Fitzpatrick spectrum for age assurance.
The probabilistic formulation maps directly onto established age-assurance frameworks, and the privacy properties of hand-based inference (no facial biometrics, low re-identification risk) make it well suited to the child-safe, data-minimising deployment contexts that regulators increasingly require.

{\small
\bibliographystyle{ieee}
\bibliography{egbib,references}
}

\end{document}